\documentclass[letterpaper]{article} 
\usepackage{aaai2026}  

\usepackage{xspace} 
\usepackage{threeparttable} 
\usepackage{enumitem} 
\usepackage{arydshln} 
\usepackage{subcaption} 

\usepackage{amsmath} 
\usepackage{amssymb} 
\usepackage{booktabs} 
\usepackage{makecell} 
\usepackage{multirow}
\usepackage{algorithm}
\usepackage{algpseudocode}

\usepackage{amsmath}   
\usepackage{amsfonts}  
\usepackage{amssymb}   
\usepackage{mathtools} 

\usepackage[most]{tcolorbox}

\newcommand{\ie}{\emph{i.e.,}\xspace}

\usepackage{times}  
\usepackage{helvet}  
\usepackage{courier}  
\usepackage[hyphens]{url}  
\usepackage{graphicx} 
\usepackage{natbib}  
\usepackage{caption} 
\usepackage{newfloat}
\usepackage{listings}
\DeclareCaptionStyle{ruled}{labelfont=normalfont,labelsep=colon,strut=off} 
\floatstyle{ruled}
\newfloat{listing}{tb}{lst}{}
\floatname{listing}{Listing}
\newcommand{\framework}{RoSA\xspace}

\newcommand{\codelink}{https://github.com/Applied-Machine-Learning-Lab/RoSA}
\title{RoSA: Enhancing Parameter-Efficient Fine-Tuning via \\ RoPE-aware Selective Adaptation in Large Language Models}
\author {
    Dayan Pan\textsuperscript{\rm 1,\rm 3},
    Jingyuan Wang\textsuperscript{\rm 1,\rm 2,\rm 3}\footnote{Corresponding authors.},
    Yilong Zhou\textsuperscript{\rm 1,\rm 3},
    Jiawei Cheng\textsuperscript{\rm 1,\rm 3,\rm 4},
    Pengyue Jia\textsuperscript{\rm 4},
    Xiangyu Zhao\textsuperscript{\rm 4}\footnotemark[1]
}
\affiliations {
    \textsuperscript{\rm 1}School of Computer Science and Engineering, Beihang University, Beijing, China\\
    \textsuperscript{\rm 2}School of Economics and Management, Beihang University, Beijing, China\\
    \textsuperscript{\rm 3}MOE Engineering Research Center of Advanced Computer Application Technology, Beihang University, China\\
    \textsuperscript{\rm 4}Department of Data Science, City University of Hong Kong, Hong Kong, China\\
    \{dayan, jywang, yilongzhou, JarvisC\}@buaa.edu.cn, jia.pengyue@my.cityu.edu.hk, xianzhao@cityu.edu.hk
}

\begin{document}

\maketitle

\begin{abstract}
Fine-tuning large language models is essential for task-specific adaptation, yet it remains computationally prohibitive. Parameter-Efficient Fine-Tuning (PEFT) methods have emerged as a solution, but current approaches typically ignore the distinct roles of model components and the heterogeneous importance across layers, thereby limiting adaptation efficiency.
Motivated by the observation that Rotary Position Embeddings (RoPE) induce critical activations in the low-frequency dimensions of attention states, we propose RoPE-aware Selective Adaptation (RoSA), a novel PEFT framework that allocates trainable parameters in a more targeted and effective manner.
RoSA comprises a RoPE-aware Attention Enhancement (RoAE) module, which selectively enhances the low-frequency components of RoPE-influenced attention states, and a Dynamic Layer Selection (DLS) strategy that adaptively identifies and updates the most critical layers based on LayerNorm gradient norms.
By combining dimension-wise enhancement with layer-wise adaptation, RoSA achieves more targeted and efficient fine-tuning.
Extensive experiments on fifteen commonsense and arithmetic benchmarks demonstrate that RoSA outperforms existing mainstream PEFT methods under comparable trainable parameters. The code is available to ease reproducibility\footnote{\codelink}.
\end{abstract}


\section{Introduction} \label{sec:intro}
\begin{figure}[t]
  \centering
  \begin{subfigure}[b]{0.47\linewidth}
    \hspace{-3px}
    \includegraphics[width=\linewidth]{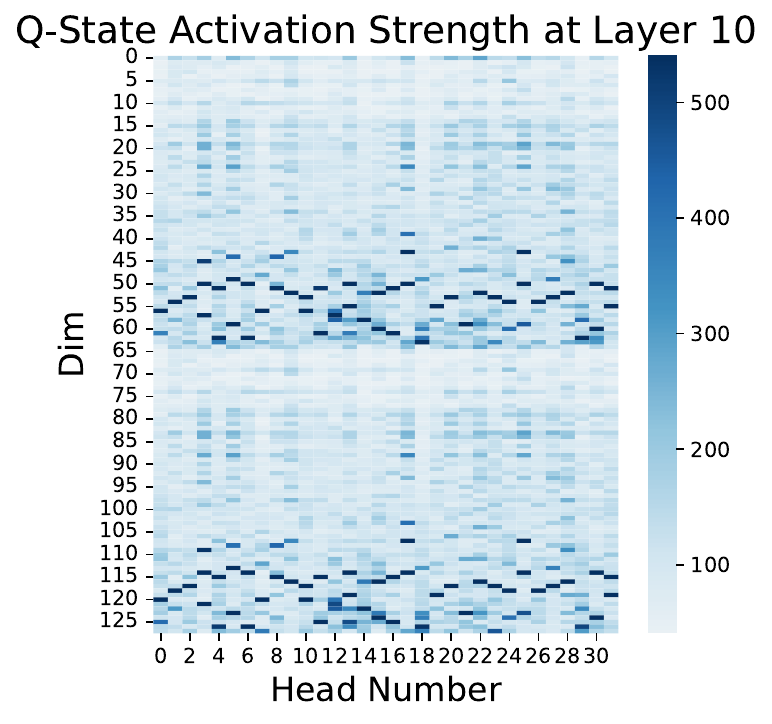}
    \caption{Across Head Dimensions}
    \label{fig:attnindim}
  \end{subfigure}
  \hfill
  \begin{subfigure}[b]{0.48\linewidth}
    \hspace{-3px}
    \includegraphics[width=\linewidth]{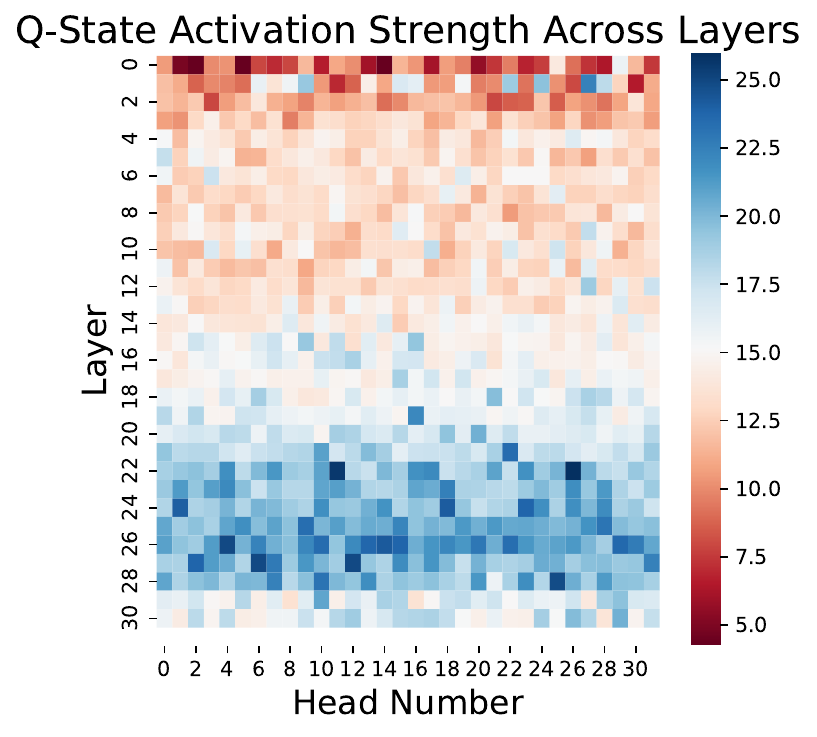}
    \caption{Across Layers}
    \label{fig:attninlayer}
  \end{subfigure}
  \caption{Q-state activation strength visualizations in LLaMA-2-7B.
We compute the average L2 norm per attention head to quantify activation strength.
Stronger activations are concentrated in high-indexed (\ie low-RoPE frequency) dimensions and vary across layers, highlighting both dimension-wise and layer-wise heterogeneity.
}
  \label{fig:hotattn}
\end{figure}
Large Language Models (LLMs) have achieved remarkable success across a wide range of natural language processing (NLP) tasks, becoming a foundational infrastructure in numerous real-world applications~\cite{cheng2025poi,yu2025bigcity}.
However, deploying these large-scale models often requires fine-tuning to align models with specific task requirements.
Traditional fine-tuning methods, such as full-parameter fine-tuning, are extremely resource-intensive, severely constraining their broader applicability. 
Consequently, exploring Parameter-Efficient Fine-Tuning~(PEFT) methods, which aim to substantially reduce fine-tuning costs without compromising model performance, has emerged as a key research focus in the LLM community~\cite{ding2023parameter, liapproximation, han2025data, liu2024full, liu2024fulltraffic}.

Recent PEFT methods typically aim to adapt LLMs to specific downstream tasks by fine-tuning only a small fraction of parameters, significantly reducing computational cost~\cite{, wang2025metalora, liu2024moe}. For example, mainstream PEFT methods such as P-tuning~\cite{liu2021p}, LoRA~\cite{hu2021lora}, DoRA~\cite{liu2024dora}, and HyCAM~\cite{pan2025contextual} introduce lightweight and trainable adaptation modules into the pre-trained model, keeping most of the original model parameters frozen. 

Despite advancements, existing PEFT methods exhibit two critical limitations:
\textbf{(1) Component-Heterogeneity Neglect:} Current methods largely neglect the intrinsic functional roles of LLM components~\cite{zhang2023adalora}. For instance, LoRA inserts low-rank matrices into the linear layers of attention and feed-forward blocks, enabling adaptation with minimal trainable parameters. However, such designs are applied uniformly across modules without analyzing their distinct functional roles.
\textbf{(2) Layer-Heterogeneity Neglect:} Existing approaches often overlook the diversity across layers.
However, LLMs capture syntax in lower layers, semantics in higher layers~\cite{voita2019bottom}.
Most PEFT methods apply uniform adaptation schemes across all layers, limiting the potential efficiency and effectiveness of parameter allocation.

Our approach is motivated by a key observation regarding LLM architectures: different components exhibit distinct roles and activation behaviors~\cite{xue2025learnable,xue2025burst, wang2025put}. 
Recent studies suggest that Feed-Forward Networks (FFN) act as repositories for storing factual knowledge, while Multi-Head Attention (MHA) modules function primarily for knowledge retrieval and contextual routing~\cite{geva2021transformer}. 
A key component within the MHA module is the Rotary Position Embedding (RoPE)~\cite{su2024roformer}, which plays a critical role in contextual understanding by encoding positional information into attention mechanisms. RoPE achieves this by applying pair-wise complex rotations to the Query (Q) and Key (K) state tensors of attention mechanism and the sinusoidal frequency increases geometrically across successive dimension pairs.

This frequency-based encoding introduces unique activation patterns.
As shown in Fig.\ref{fig:hotattn}(\subref{fig:attnindim}), there are obvious distinctions in Q-state activations across different dimensional channels.
Specifically, low-frequency components (corresponding to higher-indexed dimensions within each half of the attention states) exhibit denser and more intense activations, while high-frequency shows sparser activations.
Analyses confirm that these prominent low-frequency activations are crucial for contextual understanding~\cite{barbero2024round, jin2025massive}. 
Furthermore, Fig.\ref{fig:hotattn}(\subref{fig:attninlayer}) reveals that this activation intensity is also highly heterogeneous across different layers, suggesting their contributions are not equal. 
These findings highlight that targeting these critical low-frequency components and the varying importance across layers for fine-tuning hold significant potential for enhancing both model performance and parameter efficiency.

Building on this, we propose a novel parameter-efficient fine-tuning method called RoPE-aware Selective Adaptation (RoSA). Specifically, RoSA integrates two complementary modules:
(1) \textit{a RoPE-aware Attention Enhancement (RoAE)} module, explicitly designed to adaptively enhance the distinctive low-frequency components within query/key states influenced by the RoPE mechanism, thereby enhancing the model's contextual understanding capabilities with high parameter efficiency.
(2) a \textit{Dynamic Layer Selection~(DLS)} strategy, enabling RoSA to dynamically identify and adapt only the most critical layers during fine-tuning. Specifically, layer importance is quantified by computing the gradient norm of Layer Normalization parameters, serving as a reliable proxy for determining each layer's contribution to model performance.
By simultaneously leveraging RoPE's inherent structural characteristics and dynamically allocating fine-tuning resources to layers that matter most, RoSA substantially improves parameter efficiency and model effectiveness compared to existing PEFT techniques. The main contributions of this paper are summarized as follows:
\begin{itemize}[leftmargin=*, topsep=0pt]
    \item To our knowledge, among PEFT works, we are the first to explicitly consider the distinctive low-frequency attention components induced by RoPE and propose RoAE, a RoPE-aware PEFT module that performs targeted enhancement of these functionally key dimensions. This adaptation effectively strengthens contextual understanding capabilities in a highly parameter-efficient manner.
    \item We introduce RoSA, a comprehensive PEFT framework that combines the RoAE module with a Dynamic Layer Selection (DLS) strategy. Specifically, DLS adaptively identifies and selectively updates the most impactful layers based on gradient norms of Layer Normalization parameters. Thus, RoSA optimally allocates parameters both dimension-wise and layer-wise according to their functional importance, enhancing overall efficiency.
    \item Extensive experiments on fifteen public benchmark datasets, using three backbone models and covering commonsense and arithmetic QA tasks, demonstrate that RoSA significantly outperforms existing mainstream PEFT methods under comparable trainable parameter scales, validating both its efficiency and effectiveness.
\end{itemize}

\begin{figure*}[ht]
    \centering
    \includegraphics[width=0.81\linewidth]{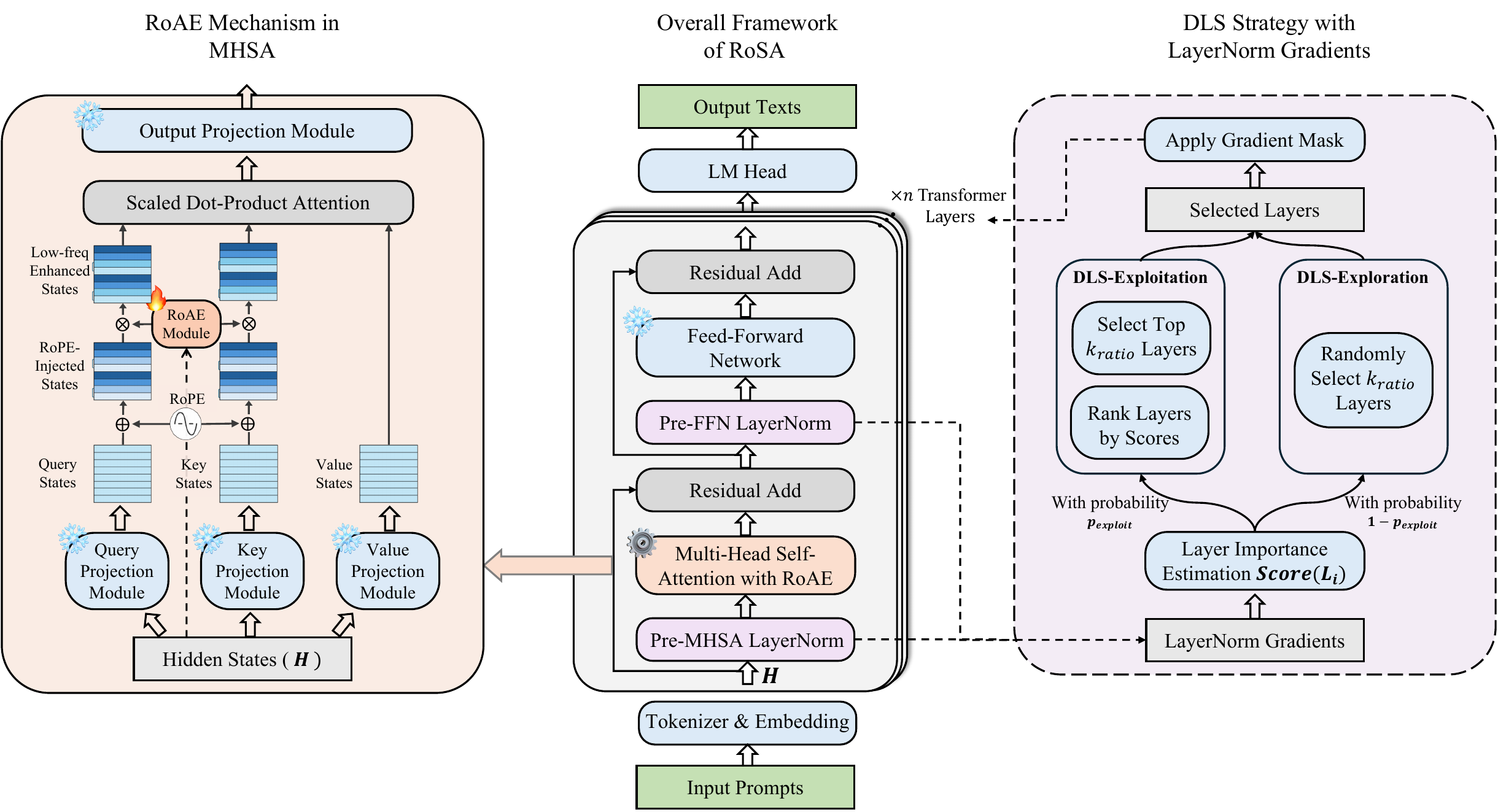}
    \caption{The architecture of RoSA. RoSA consists of two key modules: RoPE-aware Attention Enhancement (RoAE), which selectively enhances low-frequency components of RoPE-influenced Q/K states, and 
    Dynamic Layer Selection (DLS), which dynamically selects important layers for update.
    Enabling targeted, efficient adaptation both frequency-wise and layer-wise.}
    \label{fig:framework}
\end{figure*}
\section{Preliminaries} \label{sec:pre}
This section reviews the key components of LLMs and the RoPE mechanism, which form the basis of our method.
\subsection{LLM Architecture}
Modern LLMs, such as the LLaMA, are primarily built upon the decoder-only Transformer architecture~\cite{vaswani2017attention}, which has been widely adopted across diverse representation learning settings~\cite{yang2025hygmap,jiang2023self,jiang2023pdformer, han2025bridging}.
This architecture consists of a stack of identical Transformer blocks, each containing two primary components: a Multi-Head Self-Attention (MHSA) module and a Feed-Forward Network (FFN) module.
The MHSA module allows the model to weigh the importance of different tokens in the input sequence, capturing complex contextual relationships.
To incorporate crucial information about token order, which self-attention itself lacks, these models integrate positional encodings. Specifically, modern LLMs heavily adopt the Rotary Position Embedding (RoPE)~\cite{su2024roformer} as a relative positional encoding mechanism, which directly injects relative positional information into the attention computation and plays a crucial role in the model's ability to generalize over long contexts.
The FFN, typically composed of two linear layers with a non-linear activation function,
is responsible for feature transformation and is believed to be a key repository of factual and commonsense knowledge stored within the model's parameters~\cite{geva2021transformer}.
A residual connection~\cite{he2016deep} is applied around each of the two sub-modules, followed by a Layer Normalization step.
Most LLMs utilize Pre-LN for enhanced training stability, where normalization is applied directly to the input of each sub-module. In this design, LayerNorm acts as a bridge between residual stream and subsequent attention or FFN modules, modulating the information flow across modules and layers.
\subsection{Rotary Position Embedding (RoPE)}
As mentioned in the previous section, the original self-attention mechanism is inherently permutation-invariant, meaning that the order of input tokens does not affect the output.  
Therefore, an external mechanism is required to encode token positions.
While early models use additive, learned absolute position embeddings, modern LLMs widely adopt Rotary Position Embedding (RoPE)~\cite{su2024roformer} due to its effectiveness and efficiency in encoding relative positional information, especially for long sequences.

RoPE injects positional information by applying a rotational transformation directly to the Query ($q$) and Key ($k$) vectors in each attention head. 
Specifically, given a vector $\mathbf{z} \in \mathbb{R}^d$, where $d$ is even, RoPE splits it into two halves: a \textit{real} part $\mathbf{z}^{\text{real}}$ and an \textit{imaginary} part $\mathbf{z}^{\text{imag}}$, each of dimension $d/2$. Then, for each index $i$, RoPE treats $(\mathbf{z}^{\text{real}}_i, \mathbf{z}^{\text{imag}}_i)$ as a complex-valued component and applies a 2D rotation:

\begin{equation}
\text{RoPE}(\mathbf{z}^{\text{real}}_i, \mathbf{z}^{\text{imag}}_i) =
\begin{bmatrix}
\cos \theta_i & -\sin \theta_i \\
\sin \theta_i & \cos \theta_i
\end{bmatrix}
\begin{bmatrix}
\mathbf{z}^{\text{real}}_i \\
\mathbf{z}^{\text{imag}}_i
\end{bmatrix},
\end{equation}
where $\theta_i = t \cdot \omega^{-2i/d}$, $t$ is the token position index, and $\omega$ is a base frequency constant (commonly set to $10{,}000$). This operation is equivalent to applying a complex-valued sinusoidal rotation, enabling relative positional relationships to be encoded directly into the attention mechanism. Since each rotation is applied to the corresponding dimensions in the two halves of the vector, both halves share the same rotation frequency $\theta_i$. As observed in Fig.\ref{fig:hotattn}(\subref{fig:attnindim}), the activation patterns exhibit similarity, highlighting the impact of RoPE on the attention mechanism across dimensions.

As $\theta_i$ decreases geometrically with the index $i$, low-indexed dimensions encode high-frequency positional patterns, while the high-indexed dimensions encode low-frequency, smoother components.
These low-frequency components often produce stronger and denser activations, and are crucial for long-range dependency modeling.
These observations suggest that the frequency structure induced by RoPE provides a meaningful basis for improving PEFT methods.  
In this work, we explicitly target the low-frequency components of RoPE-influenced attention states, aiming to enhance parameter efficiency in a more targeted manner.

\section{Method}
In this section, we first provide an overview of the RoSA framework, then describe its two core components in detail, and finally present the overall algorithm.

\subsection{Framework Overview}
Existing PEFT methods often overlook two key aspects of LLMs: (\textit{i}) the frequency-specific structure introduced by RoPE, and (\textit{ii}) the layer-wise importance heterogeneity during adaptation.
This motivates us to design a more targeted and adaptive fine-tuning strategy.
To address these challenges, we propose RoPE-aware Selective Adaptation (RoSA). The core idea is to achieve a more targeted and efficient fine-tuning through a dual-level adaptation strategy, targeting critical low-frequency dimensions within layers and selecting the most important layers across the model.

As illustrated in Fig.\ref{fig:framework}, RoSA achieves this through two main components.
First, the RoPE-aware Attention Enhancement (RoAE) module selectively enhancing the low-frequency components of RoPE-influenced attention states, which play a critical role in contextual understanding.
Further, the Dynamic Layer Selection (DLS) module identifies and adapts the most important layers during fine-tuning based on a gradient importance metric.
By combining frequency-wise and layer-wise selective adaptation, RoSA achieves a more effective and efficient adaptation process.

\subsection{RoPE-aware Attention Enhancement (RoAE)}
Based on the observation that the low-frequency dimensions of RoPE-rotated attention states play a critical role in modeling long-range dependencies and contextual semantics~\cite{barbero2024round, jin2025massive}.
However, conventional PEFT methods do not explicitly consider this frequency structure, instead applying generic adaptations across all dimensions.
This limits their efficiency and effectiveness.
To address this, we introduce the RoPE-aware Attention Enhancement (RoAE) module, which selectively enhances the low-frequency components within the Query (Q) and Key (K) attention states in a lightweight and targeted manner.

\subsubsection{Low-Frequency Components Selection:}
Given the hidden states $\mathbf{H} \in \mathbb{R}^{b \times l \times d}$ as input to Transformer, where $b$ is the batch size, $l$ is the sequence length, and $d$ is the hidden dimension. After applying the linear projections to obtain the query and key tensors, these are reshaped into multi-head with shape $[b, h, l, d_h]$, where $h$ is the number of attention heads and $d_h = d / h$ is the dimension per head.
RoPE first splits each head vector into real $\mathbf{z}_{\text{real}}$ and imaginary $\mathbf{z}_{\text{imag}}$ halves, then applies a sinusoidal rotation to every resulting complex pair.

To extract the low-frequency components, we follow the structure of RoPE and split each head vector into two halves of size $d_h/2$. From each half, we take the last 
$(d_h \cdot r_{\text{low}}) / 2$ 
dimensions and concatenate them to form a $d_{\text{low}}$-dimensional vector, denoted as $\mathbf{z}_{\text{low}}$. Here, $r_{\text{low}} \in (0, 1)$ is a hyperparameter controlling the ratio of the targeted low-frequency components.
This extracted vector captures the critical low-frequency components of the RoPE-influenced Q/K head, serving as the target for enhancement.

\subsubsection{Adaptation Signal Generation:}
To enhance the extracted low-frequency components in a targeted way, we first generate a context-aware adaptation signal $\mathbf{S}$.
Specifically, the hidden state is passed through a trainable linear projection, $\mathbf{W}_{\text{proj}}$, followed by a non-linear activation (SiLU)~\cite{elfwing2018sigmoid} to introduce non-linearity:
\begin{equation}
\tilde{\mathbf{S}} = \text{SiLU}(\mathbf{H} \mathbf{W}_{\text{proj}}), \quad \mathbf{W}_{\text{proj}} \in \mathbb{R}^{d \times (h \cdot d_{\text{low}})},
\label{eq:roae-proj}
\end{equation}
where $\tilde{\mathbf{S}} \in \mathbb{R}^{b \times l \times (h \cdot d_{\text{low}})}$. Similarly, we then reshape the projected tensors to the multi-head shape $\mathbf{S}\in\mathbb{R}^{b \times h \times l \times d_{\text{low}}}$.

Notably, to improve parameter efficiency, the projection module $\mathbf{W}_{\text{proj}}$ is implemented using a low-rank decomposition ($\mathbf{W}_{\text{proj}} = \mathbf{B}\mathbf{A}$), adding only a small number of trainable parameters. 
Further, this design remains compatible and can be flexibly replaced by other emerging PEFT methods.

In typical settings, we use the same adaptation signal $\mathbf{S}$ for both query and key projections.
To ensure compatibility with modern architectures employing Grouped-Query Attention (GQA)~\cite{ainslie2023gqa}, where the number of query and key heads, denoted by $h_q$ and $h_k$, may differ, we apply an additional projection module to align the dimensions:
\begin{equation}
\tilde{\mathbf{S}}^{(K)} = \tilde{\mathbf{S}}^{(Q)} \cdot \mathbf{W}_{\text{GQA}}, \quad \mathbf{W}_{\text{GQA}} \in \mathbb{R}^{(h_q \cdot d_{\text{low}}) \times (h_k \cdot d_{\text{low}})},
\label{eq:roae-gqa}
\end{equation}
ensuring compatibility across varying attention configs, thereby enabling RoAE to support GQA-enabled models.

\subsubsection{Targeted Enhancement Application: }
After obtaining the adaptation signal $\mathbf{S}$, the final step is to apply it to the targeted low-frequency components.
Recall that in the previous step, we extracted the low-frequency vectors $\mathbf{z}_{\text{low}}$ of each head.
Denoting the extracted low-frequency components for all attention heads as $\mathbf{Z}\in\mathbb{R}^{b\times h\times l\times d_{\text{low}}}$, we perform the enhancement via an element-wise multiply modulation:
\begin{equation}
\mathbf{Z}^{*} = \mathbf{Z} + \mathbf{Z}\odot(\alpha\cdot\mathbf{S}),
\label{eq:roae-apply}
\end{equation}
here $\alpha$ is a scaling factor controlling the adaptation strength.

Finally, the enhanced low-frequency tensors $\mathbf{Z}^{*}$ are re-integrated into their original positions of the attention head states, replacing the corresponding low-frequency dimensions.
The attention mechanism then proceeds with these selectively enhanced query and key representations, allowing the model to better leverage RoPE's critical frequency structure for improved contextual understanding abilities.

In summary, the RoAE module introduces a targeted and efficient PEFT paradigm.
Its core innovation lies in its mechanism-aware design, which targets the critical components of RoPE-influenced attention states. 
Furthermore, the enhancement is context-aware, as the adaptation signal is dynamically generated from the input states to provide token-specific modulations.
By achieving this with high parameter efficiency and maintaining compatibility across diverse architectures, RoAE establishes a more flexible and effective method for adapting LLMs into specific tasks.

\subsection{Dynamic Layer Selection (DLS)}
While the RoAE module provides a targeted, mechanism-aware approach to adapting parameters within one layer, LLMs exhibit considerable heterogeneity across different layers, with lower layers primarily capturing syntactic features and higher layers encoding abstract semantic and contextual knowledge~\cite{voita2019bottom}.
Applying it uniformly across all layers, like common PEFT methods, overlooks the layer-wise importance heterogeneity.
To address this, we propose Dynamic Layer Selection (DLS) strategy, a method designed to dynamically select and adapt the most important layers, improving parameter utilization efficiency throughout the fine-tuning process.

\subsubsection{Layer Importance Estimation:}
The core of DLS is to accurately estimate the importance of each layer with respect to the fine-tuning objective. 
We propose to use the gradient norm of Layer Normalization (LayerNorm) parameters as an efficient proxy for this task. 
Because LayerNorm directly controls information flow between Transformer submodules and layers. A large gradient for this parameter indicates that it is necessary for the model to significantly change the output distribution of this layer to minimize the loss.

In the common-adopted Pre-LN architecture, LayerNorm modules are placed before the self-attention and before the FFN module. 
Formally, for the $i$-th Transformer layer $L_i$, its importance score is calculated by aggregating the $\text{L}_2$ norms of the gradients from the LayerNorm parameters:
\begin{equation}
\text{Score}(L_i) = \sqrt{ \| \nabla \mathbf{\Theta}_{i, \text{attn}} \|_2^2 + \| \nabla \mathbf{\Theta}_{i, \text{ffn}} \|_2^2 }
\label{eq:dls-calc}
\end{equation}
where $\mathbf{\Theta}_{i, \text{attn}}$ and $\mathbf{\Theta}_{i, \text{ffn}}$ represent the learnable parameters for the two LayerNorm modules in the $i$-th layer.
In practice, we periodically compute these importance scores for all layers, providing an informative metric to guide selection.
\subsubsection{Dynamic Selection and Gradient Masking:}
The selection procedure is activated periodically at an interval of $u$ steps after an initial warmup phase. At each activation, DLS employs a strategy that balances exploitation and exploration to choose a subset of layers for updates, specifically:
\begin{itemize}[leftmargin=*, topsep=0pt]
    \item \textbf{Exploitation:} With a high probability $p_{\text{exploit}}$, we rank all layers based on their scores and select the top-$k$ layers for training, where $k$ is determined by a predefined ratio $k_{\text{ratio}}$.
    \item \textbf{Exploration:} Conversely, with a probability of $1-p_{\text{exploit}}$, we randomly select $k$ layers to ensure that all layers have a chance to adapt, thus reducing the risk of local optima.
\end{itemize}

Once the set of selected layers $\mathcal{L_S}$ is determined, a gradient mask is applied.
Specifically, the gradients of parameters in all non-selected layers are set to 0 to prevent updating:
\begin{equation}
\nabla L_i \leftarrow \mathbf{0},\quad \text{if}\quad i \notin \mathcal{L_S}.
\label{eq:dls-mask}
\end{equation}

In summary, DLS reduces unnecessary parameter updates by dynamically identifying and adapting only the most critical layers, leading to improved efficiency and potentially superior downstream task performance. 
It is noteworthy that DLS is model-agnostic and can be easily integrated into existing PEFT pipelines. Combined with RoAE, which enables selective adaptation over important frequency components, DLS completes the RoSA framework by jointly targeting both dimension-level and layer-level adaptation.

\subsection{Overall Algorithm}
RoSA integrates the RoAE and DLS modules into the standard causal language modeling framework, where the model is trained using cross-entropy loss between predicted and target tokens. These modules operate jointly, enabling targeted adaptation both across frequency dimensions and model layers, achieving effective and efficient fine-tuning.

The full training procedure is summarized in Algorithm~\ref{alg:rosa}, 
which outlines how RoSA applies frequency-aware enhancements via RoAE and dynamically selects critical layers for update via DLS. 
Thus, RoSA optimally allocates parameters both dimension-wise and layer-wise according to their functional importance, enhancing overall efficiency.
Importantly, RoSA can be seamlessly integrated into existing PEFT frameworks or combined with other fine-tuning techniques due to its modular and adaptive design.
\begin{algorithm}[htbp]
\caption{RoPE-aware Selective Adaptation (RoSA)}
\label{alg:rosa}
\begin{algorithmic}[1]
\Require Pretrained LLM model $\mathcal{M}$, dataset $\mathcal{D}$, RoAE hyperparameters ($\alpha$, $r_{\text{low}}$), DLS hyperparameters ($k_{\text{ratio}}$, $p_{\text{exploit}}$, $u$), learning rate $\eta$, warmup steps $T_{\text{warmup}}$.
\State Initialize RoAE modules with $\alpha$ and $r_{\text{low}}$;
\State Set only RoSA-related parameters $\mathbf{\Theta}_{\text{RoSA}}$ as trainable;
\For{each training step $t$}
    \State Sample a batch of data from $\mathcal{D}$;
    \State Compute forward pass with RoAE enhanced attention states (Eq.~\ref{eq:roae-proj}-\ref{eq:roae-apply});
    \State Compute loss and perform backward pass to obtain gradients;
    \If{$t > T_{\text{warmup}}$ \textbf{and} $t$ mod $u$ == 0}
        \State Calculate layer importance $\text{Score}(L_i)$ (Eq.~\ref{eq:dls-calc});
        \State With probability $p_{\text{exploit}}$, select the top $k_{\text{ratio}}$ fraction of layers \textit{(DLS-Exploitation)}; 
        otherwise, randomly select $k_{\text{ratio}}$ fraction of layers \textit{(DLS-Exploration)};

    \EndIf
    \State Mask gradients in non-selected layers (Eq.~\ref{eq:dls-mask});
    \State Update parameters of active layers using optimizer with learning rate $\eta$;
\EndFor
\end{algorithmic}
\end{algorithm}

\begin{table*}[t]
    \centering
    \small
    \resizebox{0.96\linewidth}{!}{
        \renewcommand{\arraystretch}{0.96}
        \begin{tabular}{l|lccccccccccc}
            \toprule
            \textbf{Backbone LLM} & \textbf{Baseline} & \textbf{\# Param (\%)} & \textbf{BoolQ} & \textbf{PIQA} & \textbf{SIQA} & \textbf{ARC-C} & \textbf{ARC-E} & \textbf{OBQA} & \textbf{HellaSwag} & \textbf{WinoGrande} & \textbf{micro-avg(\%)$\uparrow$} \\
            \midrule
            \multirow{9}{*}{\textbf{Qwen 2.5 7B}}
                & LoRA&0.527         &66.9&86.8&76.7&88.2&93.9&87.2&89.7&72.2&84.3 \\ 
                & DoRA&0.546         &68.3&\underline{87.4}&77.2&\underline{89.4}&95.2&88.0&\underline{90.0}&70.4&84.9 \\ 
                & AdaLoRA&0.396      &\underline{69.7}&\underline{87.4}&\underline{77.9}&88.9&\textbf{95.7}&\underline{89.4}&\textbf{90.6}&72.6&\underline{85.6} \\ 
                & BOFT&0.023         &68.5&86.0&76.1&87.5&94.6&82.4&86.1&65.3&82.4 \\ 
                & VERA&0.018         &55.4&83.7&74.1&85.1&93.6&77.2&82.2&64.1&77.9 \\ 
                & C3A&0.665          &69.5&87.0&77.5&88.9&95.2&86.6&89.9&71.6&85.0 \\ 
                & BONE&0.291         &67.6&84.9&76.8&85.2&94.3&87.4&88.3&\textbf{77.9}&83.9 \\ 
                & LN Tuning&0.001    &62.5&86.0&73.3&85.0&93.3&77.2&80.9&62.1&78.4 \\ 
    & \framework (ours)&0.261        &\textbf{70.5}&\textbf{88.0}&\textbf{79.1}&\textbf{90.1}&\underline{95.3}&\textbf{89.6}&\textbf{90.6}&\underline{73.7}&\textbf{85.9}* \\ 
            \midrule
            \multirow{9}{*}{\textbf{Llama 3.1 8B}}
                & LoRA&0.520        &\textbf{71.7}&86.8&75.5&83.1&\underline{92.7}&82.4&\underline{88.6}&68.8&83.7 \\ 
                & DoRA&0.537        &71.5&86.9&75.8&83.2&92.5&82.2&88.5&70.0&83.8 \\ 
                & AdaLoRA&0.390     &71.1&86.2&74.7&\textbf{83.6}&92.6&82.8&87.2&\underline{70.8}&83.0 \\ 
                & BOFT&0.028        &70.5&85.5&72.4&80.0&91.9&79.0&82.4&62.5&79.7 \\ 
                & VERA&0.017        &68.8&82.9&68.4&77.6&91.4&77.4&75.2&57.4&75.2 \\ 
                & C3A&0.674         &\underline{71.6}&\textbf{87.7}&\underline{76.2}&83.1&92.6&\textbf{84.4}&88.3&70.6&\underline{83.9} \\ 
                & BONE&0.274        &64.7&78.4&74.2&72.1&86.8&78.2&81.8&70.3&77.6 \\ 
                & LN Tuning&0.003   &70.1&84.6&70.9&80.2&91.8&78.8&80.6&61.8&78.6 \\ 
    & \framework (ours)&0.329       &\textbf{71.7}&\underline{87.1}&\textbf{76.4}&\underline{83.3}&\textbf{92.8}&\underline{83.6}&\textbf{89.0}&\textbf{74.8}&\textbf{84.4}* \\ 
            \midrule            
            \multirow{9}{*}{\textbf{Gemma 2 9B}}
                & LoRA&0.581        &69.3&88.0&77.8&\textbf{88.0}&\textbf{95.5}&\underline{87.4}&89.8&\underline{77.4}&85.4 \\ 
                & DoRA&0.601        &70.0&87.3&\underline{78.1}&86.1&94.3&87.0&89.4&76.8&85.0 \\ 
                & AdaLoRA&0.437     &\underline{72.3}&\underline{88.2}&77.4&87.5&\textbf{95.5}&86.2&89.0&73.4&85.1 \\ 
                & BOFT&0.029        &65.2&83.2&72.4&81.7&91.1&75.0&80.3&62.1&77.7 \\ 
                & VERA&0.020        &65.2&79.8&66.0&73.8&85.8&61.8&70.5&56.1&70.9 \\ 
                & C3A&0.699         &70.7&87.7&77.7&86.9&\underline{94.5}&86.8&\textbf{90.4}&75.3&\underline{85.5} \\ 
                & BONE&0.319        &60.3&75.3&66.3&69.0&83.7&74.0&67.3&64.3&68.7 \\ 
                & LN Tuning&0.007   &61.2&78.1&66.1&73.2&85.0&65.0&71.9&55.1&70.7 \\ 
        & \framework (ours)&0.363   &\textbf{74.0}&\textbf{88.3}&\textbf{78.5}&\underline{87.8}&\textbf{95.5}&\textbf{87.8}&\underline{90.0}&\textbf{77.5}&\textbf{86.2}* \\ 
            \bottomrule
        \end{tabular}
    }
    \caption{Performance comparison of RoSA and baseline methods on the Commonsense QA task across three backbone LLMs. 
    \textbf{{\large *}} indicates the statistically significant improvements (\ie two-sided t-test with $p<0.05$) over the best baseline. 
    RoSA consistently achieves the highest average performance under comparable parameter budgets.}
    \label{tab:main_common}
\end{table*}

\begin{table}[t]
    \centering
    \small
    \resizebox{1\linewidth}{!}{
        \renewcommand{\arraystretch}{1}
        \begin{tabular}{lcccc}
            \toprule
            \textbf{Baseline} & \textbf{Qwen2.5 0.5B} & \textbf{Qwen2.5 1.5B} & \textbf{Qwen2.5 3B} & \textbf{Qwen2.5 7B} \\
            \midrule
                AdaLoRA           &\underline{53.5}&\underline{75.1}&81.1&\underline{85.6} \\
                C3A               &53.1&74.9&\underline{81.2}&85.0 \\
                \framework (ours) &\textbf{53.7}&\textbf{75.5}&\textbf{82.0}&\textbf{85.9} \\
            \bottomrule
        \end{tabular}
    }
    \caption{Average Commonsense QA accuracy of RoSA, AdaLoRA, and C3A on varying sizes Qwen2.5 (0.5 to 7B). 
    \label{tab:scale}
    }    
\end{table}
\section{Experiments}
To comprehensively evaluate the performance of our proposed RoSA, we conduct extensive experiments guided by the following key research questions (RQs):

\begin{itemize}[leftmargin=*]
    \item \textbf{RQ1:} How does RoSA perform compared to state-of-the-art PEFT methods across different backbone LLMs and downstream tasks?
    \item \textbf{RQ2:} How does RoSA demonstrate scalability performance with backbone LLMs of different parameter sizes?
    \item \textbf{RQ3:} What are the contributions of each component within RoSA (RoAE and DLS) to its overall performance?
    \item \textbf{RQ4:} How do RoSA's key hyperparameters affect its overall performance?
\end{itemize}

We first introduce the experimental setup and then systematically address each of the above research questions.

\subsection{Experimental Setup}
\subsubsection{Datasets}
We follow LLM-Adapters~\cite{hu2023llm} and evaluate RoSA on two distinct tasks: Commonsense QA and Arithmetic QA. 
Specifically, we fine-tune models using \texttt{Commonsense15K} and  \texttt{Math10K}, which are constructed from multiple data sources.
For the \textit{Commonsense} task, we evaluate on eight diverse benchmarks
: BoolQ, PIQA, SIQA, ARC-Challenge, ARC-Easy, OBQA, HellaSwag, and WinoGrande.
Further, we assess performance of the \textit{Arithmetic} task on seven benchmarks: MultiArith, GSM8K, AddSub, AQuA, SingleEq, SVAMP, and MAWPS.
We report accuracy on each benchmark as the evaluation metric.

\subsubsection{Backbone Models}
We select three powerful and widely-used LLMs as backbone models to validate the generalization of RoSA: Qwen2.5-7B~\cite{bai2023qwen}, Llama-3.1-8B~\cite{dubey2024llama}, and Gemma2-9B~\cite{team2024gemma}.

\subsubsection{Baseline Methods}
We evaluate our approach against a comprehensive set of recent and diverse PEFT methods. Specifically, we compare several low-rank methods and their variants, including the basic \textbf{LoRA}~\cite{hu2021lora}, its weights decomposing successor \textbf{DoRA}~\cite{liu2024dora}, dynamically rank-allocating \textbf{AdaLoRA}~\cite{zhang2023adalora}, and shared low-rank matrices \textbf{VERA}~\cite{kopiczko2023vera}.
Methods leveraging more complex structured matrices, such as the orthogonality-enforcing \textbf{BOFT}~\cite{liu2023parameter}, the circular-convolution-based \textbf{C3A}~\cite{chen2024parameter}, and the block-affine-transformation-based \textbf{BONE}~\cite{kang2024balancing} are also introduced. Finally, a simple and effective method \textbf{LN Tuning}~\cite{zhao2023tuning} is included, which only tunes the model's Layer Normalization parameters.

\subsubsection{Implementation Details}
All experiments are conducted on NVIDIA GeForce RTX 3090 with PyTorch and Transformers.
We use an AdamW optimizer with a learning rate of 1e-3.
Hyperparameters are as follows: low-freq dimension ratio $r_{\text{low}}$: 0.25, scaling factor $\alpha$: 0.1, low-rank projection dimension: 128, layer selection ratio $k_{\text{ratio}}$: 0.5,  selection interval $u$: 40 steps and exploitation probability $p_{\text{exploit}}$: 0.8.

\subsection{Overall Performance (RQ1, 2)}
To answer RQ1, we compare RoSA against all baselines on two distinct tasks: Commonsense and Arithmetic QA. 
The results are summarized in Table \ref{tab:main_common} and Table \ref{tab:main_arith}, respectively.

As shown in Table \ref{tab:main_common}, RoSA consistently achieves the best performance across all three backbone models, maintaining relatively low trainable parameters. This confirms that the low-frequency components introduced by RoPE play a crucial role in improving the model's contextual understanding.
Among LoRA variants, AdaLoRA's dynamic rank allocation yields better performance, aligning with the principles of dynamic selection of DLS module. Methods like C3A, which employ novel adapter designs, also show competitive results, highlighting the potential of more complex structured matrices for improving parameter efficiency.
Additionally, LN Tuning, a simple and effective method, performs well with minimal trainable parameters, further supporting the use of LayerNorm as an importance proxy in DLS.

To validate RoSA's capabilities, we also conduct a focused comparison on the Arithmetic QA task, specifically using the Qwen2.5-7B model due to space constraints. The results, summarized in Table \ref{tab:main_arith}, are consistent with those observed in the Commonsense task, where RoSA still achieves the best performance among all methods.

To further answer RQ2, we investigate how RoSA's performance scales with model size. We evaluate four Qwen2.5 variants (0.5B, 1.5B, 3B, and 7B) on the Commonsense QA task, comparing against two strong baselines, AdaLoRA and C3A. As shown in Table~\ref{tab:scale}, all methods improve with larger models, but RoSA consistently maintains a clear advantage across scales, highlighting its robustness and scalability.

\begin{table*}[t]
    \centering
    \small
    \resizebox{0.94\linewidth}{!}{
        \renewcommand{\arraystretch}{0.92}
        \begin{tabular}{lcccccccccc}
            \toprule
             \textbf{Baseline} & \textbf{\# Param (\%)} & \textbf{MultiArith} & \textbf{GSM8K} & \textbf{AddSub} & \textbf{AQuA} & \textbf{SingleEq} & \textbf{SVAMP} & \textbf{MAWPS} & \textbf{micro-avg(\%)$\uparrow$} \\
            \midrule
                LoRA&0.527            &93.0&68.7&88.8&33.8&\underline{88.9}&79.2&88.2&77.7 \\ 
                DoRA&0.546            &92.3&\underline{70.0}&88.6&34.6&88.5&79.6&87.3&78.1 \\ 
                AdaLoRA&0.396         &90.0&68.8&85.3&33.8&85.6&78.9&84.0&76.3 \\ 
                BOFT&0.023            &89.6&67.8&82.5&31.1&86.2&75.2&80.2&74.6 \\ 
                VERA&0.018            &72.5&63.7&80.7&31.1&80.3&74.2&83.1&70.0 \\ 
                C3A&0.665             &\textbf{95.3}&67.1&\underline{90.3}&\textbf{35.4}&\textbf{90.1}&\underline{82.1}&\underline{89.4}&\underline{78.7} \\ 
                BONE&0.291            &92.8&66.6&89.6&33.4&88.3&\underline{82.1}&89.0&77.8 \\ 
                LN Tuning&0.001       &79.6&63.6&72.1&34.2&75.3&68.1&70.1&67.7 \\ 
             \framework (ours)&0.261   &\underline{94.3}&\textbf{71.3}&\textbf{92.1}&\underline{35.0}&\textbf{90.1}&\textbf{82.2}&\textbf{92.0}&\textbf{80.1}* \\ 
            \bottomrule
        \end{tabular}
    }
    \caption{Evaluation of RoSA and baseline methods on the Arithmetic QA task using the Qwen2.5-7B model. RoSA achieves the highest average accuracy across all benchmarks, demonstrating its generalization to mathematical tasks.}
    \label{tab:main_arith}
\end{table*}

\subsection{Ablation and Hyperparameter Analysis~(RQ3, 4)}
We then perform ablation and hyperparameter studies to analyze RoSA components and sensitivity to hyperparameters.
All results in this section are reported as average performance on the Commonsense QA task with Qwen2.5-7B.
\subsubsection{Ablation Study: } 
We first conduct an ablation study comparing the full RoSA framework against several variants to evaluate the contributions of its components, as shown in Table~\ref{tab:abla}.
The full \textbf{RoSA} model includes both RoAE and DLS. We first examine the \textbf{RoSA-RoAEonly} variant by disabling DLS for evaluating the impact of layer selection.
We further investigate several RoAE replacement and modification variants, all retaining DLS:
(i) \textbf{RoSA-RoAE0.5}, which sets the low-freq dimension ratio $r_{\text{low}}$ to 0.5 while keeping all other settings unchanged,
(ii) \textbf{RoSA-Lr128}, which applies standard LoRA on Q/K with all other configs identical to RoSA, and 
(iii) \textbf{RoSA-Lr64}, which uses LoRA with a similar number of trainable parameters as RoSA.
These variants also provide an implicit analysis of the effect of $r_{\text{low}}$, allowing us to compare targeted adaptation on varying frequency ranges.
Overall, the results indicate that each component of RoSA contributes to performance, and focusing adaptation on a compact low-frequency subspace is more effective.

\subsubsection{Sensitivity of DLS: }

\begin{figure}[tb]
\vspace{-3pt}
  \centering
  \resizebox{0.98\linewidth}{!}{

    \begin{minipage}[t]{.45\columnwidth}
      \centering
      \captionsetup{font=small}
      \vspace{-0.1pt}
        \renewcommand{\arraystretch}{0.95}
        \resizebox{.9\linewidth}{!}{
          \begin{tabular}{lc}
            \toprule
            \textbf{Variant} & \textbf{micro-avg$\uparrow$}\\
            \midrule
            RoSA   & \textbf{85.9}\\
            \makecell[l]{RoSA-RoAEonly\\ (w/o DLS)}      & 84.8\\
            \makecell[l]{RoSA-RoAE0.5\\ (w/ DLS \& RoAE)} & 85.6\\
            \makecell[l]{RoSA-Lr128\\ (w/ DLS, w/o RoAE)} & 83.9\\
            \makecell[l]{RoSA-Lr64\\ (w/ DLS, w/o RoAE)}  & 80.7\\
            \bottomrule
          \end{tabular}}
      \captionof{table}{Ablation results of RoSA on Commonsense task using Qwen2.5-7B.}
      \label{tab:abla}
    \end{minipage}
    \hfill
    \begin{minipage}[t]{.45\columnwidth}
      \vspace{0.1pt}
      \centering
        \includegraphics[width=\linewidth]{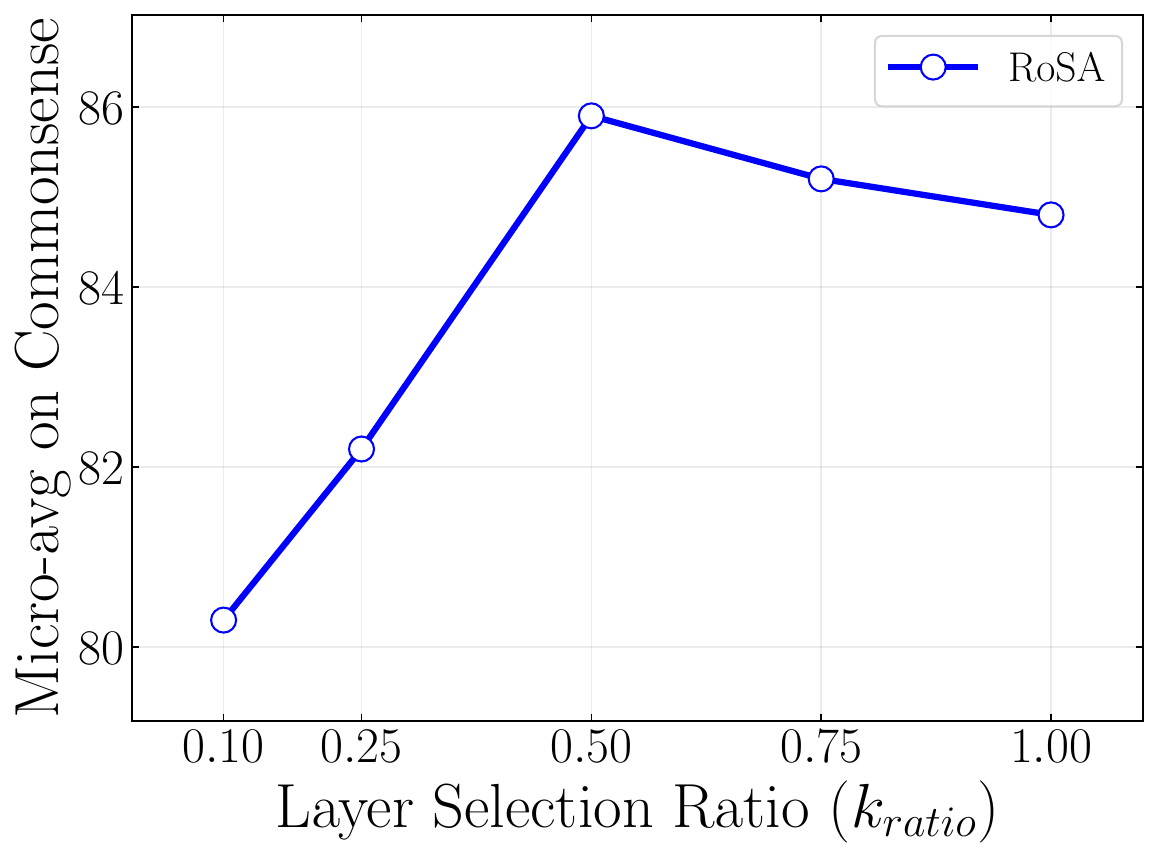}
      \captionsetup{font=small}
      \captionof{figure}{Impact of layer selection ratio $k_{\text{ratio}}$.}
      \label{fig:sens_dls}
    \end{minipage}
  }
    \vspace{-9px}
\end{figure}

To further evaluate the DLS module, we analyze the sensitivity of the layer selection ratio $k_{\text{ratio}}$, which controls the proportion of layers updated during fine-tuning. We vary $k_{\text{ratio}}$ over a range of values. As summarized in Fig.~\ref{fig:sens_dls}, RoSA performs best when $k_{\text{ratio}} \approx 0.5$. Increasing this ratio slightly degrades performance, suggesting that selectively updating fewer layers leads to more efficient optimization and enhances overall model performance.

\section{Related Work}

\subsection{Parameter-Efficient Fine-Tuning}
Parameter-Efficient Fine-Tuning (PEFT) aims to adapt LLMs to downstream tasks by tuning only a small subset of parameters, significantly reducing computational and memory costs.
Adapter-based methods insert small trainable modules, enabling effective task adaptation with minimal parameters~\cite{houlsby2019parameter}.
Low-rank methods like LoRA~\cite{hu2021lora} and its variants, including DoRA~\cite{liu2024dora}, AdaLoRA~\cite{zhang2023adalora}, and VERA~\cite{kopiczko2023vera}, inject trainable low-rank matrices into pretrained weights to achieve efficient adaptation.
Advanced structured-matrix methods, such as C3A~\cite{chen2024parameter} and BONE~\cite{kang2024balancing}, introduce circular convolution or block affine into PEFT, further enhancing parameter efficiency through structured constraints.
These efforts complement broader work on model efficiency, including compression and distillation techniques~\cite{wang2023large}, as well as domain-specific sequence modeling frameworks~\cite{wang2025gtg} and efficient decision-making systems~\cite{cong2021alphaportfolio}.
However, most existing methods apply adaptation uniformly across model components, often neglecting their distinct functional roles.

\subsection{Analysis of LLM Internals}
Understanding the internal mechanics of LLMs is a growing research area that provides crucial insights for developing more principled and efficient methods.
Early research shows that each FFN can be seen as a key-value memory~\cite{geva2021transformer}.
Recent work provides evidence that attention mechanisms are crucial for retrieving relevant context and enabling dynamic reasoning~\cite{dong2025attention,zhang2025process}, whereas the FFN layers are responsible for memorizing task-specific or factual content.
RoPE in particular has been discussed in recent studies, inducing strong and dense activations in the low-frequency dimensions of attention states, and these activations are crucial for the LLMs' contextual understanding capabilities~\cite{jin2025massive,barbero2024round}. 
The frequency-structured behavior of attention has also been examined in wavelet-based or efficient attention training frameworks~\cite{wang2023wavelet,fu2025sliding}.
Meanwhile, analyses of layer-wise behavior reveal that not all layers are equally important~\cite{belinkov2018evaluating}, a trend also echoed in broader structure-aware neural modeling literature~\cite{ji2022stden,hettige2024airphynet,wang2022traffic}. 
These findings underscore that different submodules contribute unique and complementary functions in LLMs, motivating our RoSA method.
\section{Conclusion}
In this work, we introduce RoPE-aware Selective Adaptation (RoSA), a novel PEFT framework for LLMs.
RoSA explicitly leverages the frequency structure induced by RoPE by introducing a RoPE-aware Attention Enhancement (RoAE) module, which selectively enhances low-frequency attention components. Alongside, the Dynamic Layer Selection (DLS) strategy dynamically identifies and updates the most important layers based on LayerNorm gradients.
This dual-level design enables more effective and targeted use of trainable parameters both within and across layers.
Extensive experiments on fifteen commonsense and arithmetic QA datasets, covering multiple LLM families and model sizes, demonstrate that RoSA consistently outperforms baseline PEFT methods under comparable trainable parameters.

\section{Acknowledgments}
Jingyuan Wang's work was partially supported by the National Natural Science Foundation of China (No. 72171013, 72222022, 72242101), and the Fundamental Research Funds for the Central Universities (JKF-2025017226182).
This research was partially supported by National Natural Science Foundation of China (No.62502404), Hong Kong Research Grants Council (Research Impact Fund No.R1015-23, Collaborative Research Fund No.C1043-24GF, General Research Fund No.11218325), Institute of Digital Medicine of City University of Hong Kong (No.9229503), Huawei (Huawei Innovation Research Program), Tencent (CCF-Tencent Open Fund, Tencent Rhino-Bird Focused Research Program), Alibaba (CCF-Alimama Tech Kangaroo Fund No. 2024002), Didi (CCF-Didi Gaia Scholars Research Fund), Kuaishou, and Bytedance.
\bibliography{ref}

\end{document}